\documentclass[letterpaper, 10 pt, conference]{ieeeconf}  

\IEEEoverridecommandlockouts                              

\usepackage{cite}
\usepackage{amsmath,amssymb,amsfonts}
\usepackage{algorithmic}
\usepackage{graphicx}
\usepackage{textcomp}

\usepackage{tabularx}
\usepackage{array,tabularx}
\usepackage{multicol} 
\usepackage{diagbox}
\usepackage{multirow}
\usepackage{hyperref}
\usepackage{booktabs}
\usepackage{bm}

\title{\LARGE \bf
Fast State-of-Health Estimation Method for Lithium-ion Battery \\ using Sparse Identification of Nonlinear Dynamics
}

\author{Jayden Dongwoo Lee$^{1}$, Donghoon Seo$^{2}$, Jongho Shin$^{2}$ and Hyochoong Bang$^{1}$
\thanks{*Jayden Dongwoo Lee and Donghoon Seo contributed equally to this work.}
\thanks{$^{1}$Jayden Dongwoo Lee and Hyochoong Bang are with the Korea Advanced Institute of Science and Technology, 373-1 Kusong-Dong, Yusong-Gu, Daejon, 305-701, Korea
        {\tt\small cin6474@kaist.ac.kr; hcbang@kaist.ac.kr}}%
\thanks{$^{2}$Donghoon Seo and Jongho Shin are with the Chungbuk National University, 1, Chungdae-ro, Seowon-gu, Cheongju-si, Chungcheongbuk-do, Korea
        {\tt\small cryface8756@chungbuk.ac.kr; jshin@chungbuk.ac.kr}}%
}

\begin{document}

\maketitle
\thispagestyle{empty}
\pagestyle{empty}

\begin{abstract}

Lithium-ion batteries (LIBs) are utilized as a major energy source in various fields because of their high energy density and long lifespan. During repeated charging and discharging, the degradation of LIBs, which reduces their maximum power output and operating time, is a pivotal issue.
This degradation can affect not only battery performance but also safety of the system. Therefore, it is essential to accurately estimate the state-of-health (SOH) of the battery in real time. To address this problem, we propose a fast SOH estimation method that utilizes the sparse model identification algorithm (SINDy) for nonlinear dynamics. SINDy can discover the governing equations of target systems with low data assuming that few functions have the dominant characteristic of the system. To decide the state of degradation model, correlation analysis is suggested. Using SINDy and correlation analysis, we can obtain the data-driven SOH model to improve the interpretability of the system. To validate the feasibility of the proposed method, the estimation performance of the SOH and the computation time are evaluated by comparing it with various machine learning algorithms.

\end{abstract}

\section{INTRODUCTION}
\label{sec:introduction}
Lithium-ion batteries (LIBs) are a major energy source used in various fields such as transportation, aviation, and military \cite{Olabi}. They have a high energy density and low self-discharge rate, providing high power output and long operation time \cite{Kebede}. However, due to irreversible chemical reactions, the internal resistance of LIBs increases with repeated use, especially under abnormal conditions such as shock or misuse \cite{Xiong}. An increase in internal resistance indicates an aging of the battery, which means that the maximum usable capacity has decreased. A battery with a reduced maximum capacity will have lower performance, including maximum power output and operating time, and an increased risk of safety incidents \cite{Wang}.
Therefore, methods to ensure the safety and reliability of systems using LIBs are essential. Related research has been expanded to include state of charge (SOC), state of health (SOH), and remaining useful life (RUL) \cite{Hu}, \cite{SOC}. At this time, these variables are derived based on the maximum available capacity $C_{\text{chargeable}}$ and are calculated as

\begin{align}
    \text{SOC}(\%)=\frac{C_{\text{chargeable}} - C_{\text{discharge}}}{C_{\text{chargeable}}} \times 100,
    \label{SOC}
\end{align}
\begin{align}
    \text{SOH}(\%)=\frac{C_{\text{chargeable}}}{C_{\text{nominal}}} \times 100,
    \label{SOH}
\end{align}
\begin{align}
    \text{RUL} = \max_{i \in (1, k)} \left({C_{\text{chargeable}, i}} \geq C_{\text{Threshold}} \right),
    \label{RUL}
\end{align}
where ${C_{\text{nominal}}}$ is the nominal capacity of battery and $C_{\text{Threshold}}$ is the threshold capacity of battery.  

Therefore, it is essential to identify the maximum chargeable capacity to diagnose the system. Most studies have used data obtained in reliable environments using specialized equipment, such as impedance spectroscopy \cite{Choi} and safety and useful life experiments \cite{Chen}, which are difficult to apply to real-world conditions. To improve this, many investigations have been performed to identify the maximum usable capacity in real-world environments by utilizing raw sensor data such as voltage, current, and temperature.

Jiang et al. \cite{Jiang} utilized an incremental capacity analysis (ICA) and Kalman filter to identify the maximum available capacity. Schaltz et al. \cite{Schaltz} studied an incremental capacity-based estimation method for battery packs. The study showed similar performance to the results for cells, demonstrating the validity of the method. In \cite{Bian}, a new method was proposed that combines open circuit model and ICA to account for a noise in the measured data. In \cite{Guo}, differential voltage analysis (DVA) was used to identify singularities of degradation in batteries and utilize them to estimate a capacity. To improve DVA performance degradation caused by voltage measurement noise, Zhu et al. \cite{Zhu} proposed a capacity estimation method using a Kalman filter and a particle filter. In \cite{Pan}, a capacity estimation algorithm was proposed combining ICA and DVA to enhance the performance of SOH estimation. ICA and DVA are very effective in analyzing aging phenomena and have satisfactory estimation performance, as they extract voltage variations due to battery degradation. However, the application of these methods requires data acquired at very low current magnitudes.

To overcome this limitation, machine learning methods that utilize large data sets have been studied. Feng et al. \cite{Feng} extracted features from the partial charging data and performed regression analysis using a support vector machine (SVM). In \cite{Zhang}, a combined least squares support vector machine (LSSVM) and an error compensation model were proposed to efficiently map nonlinear systems with data sparsity and degradation. In \cite{Lyu} and \cite{Chen3}, SOH estimation methods using relevance vector machine (RVM) were proposed to capture and utilize the relationship between incremental capacity data and differential pressure data, respectively, and showed that they are effective even for single data. In \cite{Reddy}, Gaussian process regression (GPR) was used to estimate SOH using valid input variables that are selected by principal component analysis (PCA). However, these machine learning methods require large data to train and a long computation time to estimate SOH that hinders the use of a battery management system in real time.  

Recently, to address the challenge of these machine learning methods, a sparse identification of nonlinear dynamics (SINDy) is proposed \cite{sindy1}. The SINDy can discover the governing equation of
system with low data assuming that few functions have the dominant characteristic of the system. Leveraging this feature, SINDy is able to provide a meaningful representation of the system relative to deep neural networks and Gaussian process method. Due to these advantages, SINDy has been used to model systems in various fields such as quadrotors \cite{sindy2}, soft robots \cite{sindy3}, and chemical processes \cite{sindy4}. In addition, novel research has been conducted on the
data-driven discovery of lithium-ion battery state of charge dynamics using SINDy \cite{sindy5}. However, this method assumes that we can know the maximum available capacity at every time, but it is not possible in real time. Motivated by this idea and to address issues about fast estimation of SOH, we propose a data-driven SOH estimator for LIBs to rapidly diagnose the health of battery. 

The paper makes the following contributions: (1) data-driven modeling method can provide a meaningful representation for a degradation of LIB that is difficult to model because of the electrochemical reaction (2) A correlation analysis for input variables is proposed to decide the state of data-driven model (3) SINDy-based SOH estimator is suggested to obtain an accurate and fast SOH value in real time. To the best of our knowledge, the proposed method is the first application of real-time battery SOH estimator using SINDy.

This paper is structured as follows: Section \ref{sec2} demonstrates how to obtain a dataset and extract features to determine the
state of SINDy. Section \ref{sec3} introduces a SINDy-based SOH estimator. In Section \ref{sec4}, the simulation results and discussion are presented. Section \ref{sec5} provides a conclusion and future work.

\section{Data Preparation}
\label{sec2}
This chapter describes the datasets that the research group obtained from the performance experiments and the process of preparing the input and output data.

\subsection{LIB Experimental Dataset}
The dataset utilized in this study consists of time series data (voltage, current) acquired through charge and discharge performance experiments. 8 LIBs with a nominal voltage of 3.7V and a capacity of 2Ah were used in the experiments.

\begin{center}
\begin{table}[h!]
\caption{Experimental condition in Lab}
\centering
\begin{tabular}{|cc|cc|}
\hline
\multicolumn{1}{|c}{} & \multicolumn{1}{c|}{} & \multicolumn{2}{c|}{Process} \\ \cline{3-4}
\multicolumn{2}{|c|}{} & \multicolumn{1}{c|}{Reference} & \multicolumn{1}{c|}{Aging} \\ \hline

\multicolumn{2}{|c|}{\begin{tabular}[c]{@{}c@{}}Charge\end{tabular}} & \multicolumn{2}{c|}{\begin{tabular}[c]{@{}c@{}}CC : 1.25A charge, 4.2V cut-off\end{tabular}}  \\
\multicolumn{2}{|c|}{\begin{tabular}[c]{@{}c@{}}(CC-CV)\end{tabular}} & \multicolumn{2}{c|}{\begin{tabular}[c]{@{}c@{}}CV : 4.2V charge, 125mA cut-off\end{tabular}}  \\ \hline

\multicolumn{2}{|c|}{\begin{tabular}[c]{@{}c@{}}Discharge\end{tabular}} & \multicolumn{1}{c|}{1.25A Discharge}  & \multicolumn{1}{c|}{rand(1.25A~5A)}  \\
\multicolumn{2}{|c|}{\begin{tabular}[c]{@{}c@{}}(CC)\end{tabular}} & \multicolumn{1}{c|}{3.2V cut-off}  & \multicolumn{1}{c|}{max 5 minute}  \\ \hline

\end{tabular}
\label{Lab}
\end{table}
\end{center}

Table \ref{Lab} shows the experimental process, and the charging and discharging process performed during the experiment is performed by the CC-CV charging protocol method, which integrates constant current (CC) and constant voltage (CV).

The reference process (RP) is used to determine the maximum rechargeable capacity of an aged LIB during an experiment and involves two full cycles. The first charge and discharge are used to initialize the residual capacity, and the second charge and discharge are used to determine the maximum available capacity. The capacity is derived based on the current accumulation method as follows:
\begin{align}
    C_{\text{chargeable}} = \int_{0}^{t_{\text{CC}}}{I_{\text{CC}}(\tau)d\tau} + \int_{0}^{t_{\text{CV}}}{I_{\text{CV}}(\tau)d\tau},
\label{eq_2}
\end{align}
where $t_{\text{CC}}$ and $t_{\text{CV}}$ are the time spent in CC and CV processes, respectively, and $I_{\text{CC}}$ and $I_{\text{CV}}$ are the current magnitude input to CC and CV processes, respectively.

The aging process (AP) is performed based on a randomized current profile to simulate the various behaviors of LIBs to induce degradation. In this process, a single charge is followed by up to 20 discharges to induce a full discharge. In addition, the same charging protocol as that used in the RP process is used to identify the charged capacity during degradation.

\begin{figure}[h]%
\centering
\includegraphics[width=40mm,clip]{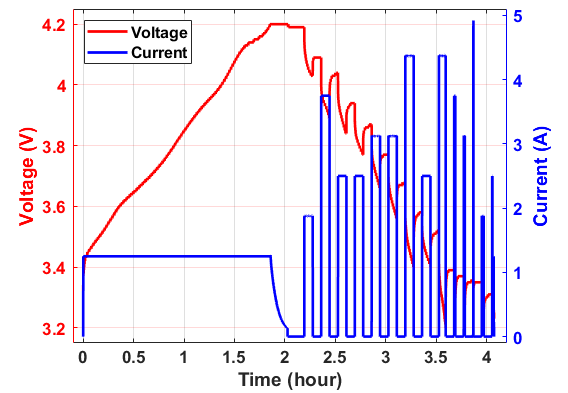}
\includegraphics[width=40mm,clip]{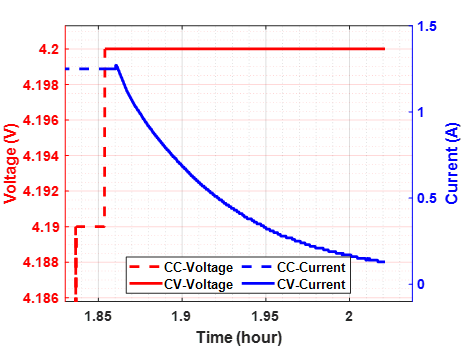}
  \caption{Aging process (L) and POV on constant voltage charging (R) in experiment.}
  \label{history_V}%
\end{figure}%

\subsection{Extraction of Inspection Data}
Unlike discharging, the charging considered in this study is carried out in a controlled environment, resulting in fewer rapid changes and therefore more reliable data. In particular, time-series data of voltage and current of battery recorded over time are useful for tracking and analyzing changes in battery health. However, the starting voltage is not consistent from one charging point to another, and the data can be inconsistent depending on the remaining capacity.

To ensure data consistency, this study extracts constant voltage charging intervals during the entire charging process and analyzes them. Constant voltage charging occurs after the battery voltage reaches its maximum, so the starting conditions are always the same regardless of the starting voltage and residual capacity.

\subsection{Feature Extraction and Correlation Analysis}
The SOH of battery is determined by multiple factors, it is difficult to measure with specific measurements. Therefore, we performed a correlation analysis using variables that may be related to the SOH of the battery. Through this analysis, the variables to be used for data-driven modeling are determined.

\subsubsection{Output data}
The SOH is derived from the maximum charge capacity, which is derived from the charging process, as mentioned in the previous section.

\begin{figure}[h]%
\centering
\includegraphics[width=75mm,clip]{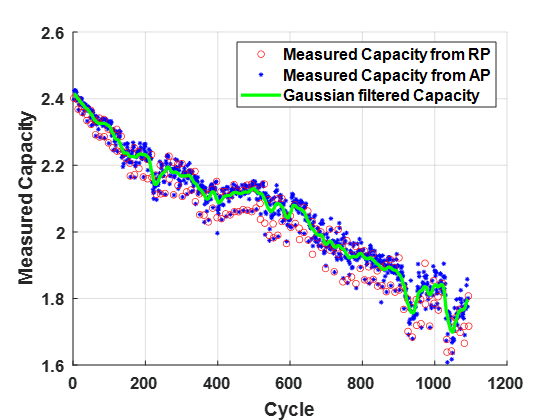}
  \caption{Filtered capacity from experimental data.}
  \label{GF}%
\end{figure}

However, the capacities measured during the RP and AP processes may differ from the maximum chargeable capacity due to various uncertainty factors such as recovery effects and residual stresses. Therefore, in this study, the capacities obtained during the RP and AP processes were combined and pre-processed by applying a Gaussian filter \cite{Korkmaz}, and the results are shown in Fig. \ref{GF}. The pre-processed capacity is converted to SOH using (\ref{SOH}).

\subsubsection{Input data}
The input data is generated from constant voltage charge data. Since constant voltage charging data has very little variation in voltage, features must be derived from variation in current to effectively generate input data.

Therefore, in this study, we used four statistical variables (mean, standard deviation, kurtosis, and skewness) that can express the distribution of the data for effective singularity capture from time series current data \cite{Seo}. In particular, kurtosis and skewness refer to the degree of pointedness and asymmetry, respectively, and are expressed as

\begin{itemize}
    \item Kurtosis ($Kur$)
    \begin{align}
    Kur = \frac{n}{(n-1)(n-2)} \sum_{i=1}^{n} \bigg(\frac{x_i - \mu}{\sigma}\bigg)^{4} - 3,
    \end{align}
    \item Skewness ($Skew$)
    \begin{align}
    Skew = \frac{1}{n-1} \sum_{i=1}^{n} \bigg(\frac{x_i - \mu}{\sigma}\bigg)^{3} ,
    \end{align}
\end{itemize}
where $n$ represents the length of the data, $x_i$ represents the value of the variable over time, and $\mu$ and $\sigma$ represent the mean and standard deviation of the input data.

\begin{figure}[h]%
\centering
\includegraphics[width=75mm,clip]{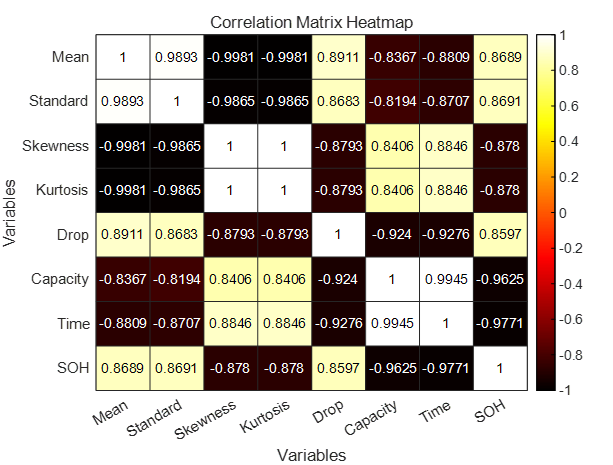}
  \caption{Correlation analysis for input variables.}
  \label{Corr}%
\end{figure}

In addition, a total of 7 input variables including constant voltage charging capacity $C_{\text{CV}}$, duration $T$, and current drop $\Delta I$, were used to find suitable input data for modeling.
Figure \ref{Corr} shows the correlation of SOH with the selected input variables. Since all variables have a correlation of 0.8 or higher, it was decided to use them for modeling.

\section{SINDy-based SOH Estimation}
\label{sec3}
\subsection{Sparse Identification of Nonlinear Dynamics}


\begin{figure}[h!]
    \centering
    \includegraphics[width=60mm,clip]{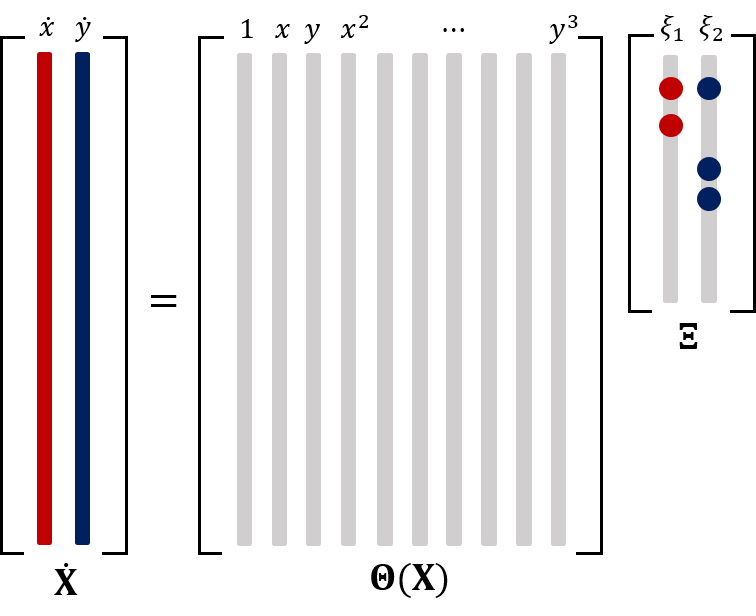}
    \caption[Concept of SINDy]{Concept of SINDy.}
    \label{fig:Concept of SINDy}
\end{figure}
Due to the development of machine learning techniques and computer computations, research on system modeling using machine learning has been extensively applied in various fields. However, traditional machine learning techniques require a large amount of data to learn the system and have problems with overfitting. Additionally, they do not provide meaningful governing equations of the system. To address these issues, SINDy method is proposed. SINDy is a technique that selects the basis that significantly affects the modeling from the entire candidate functions through an L1 penalty term. Unlike traditional machine learning techniques, it has the advantage of learning the system model with less overfitting with a small amount of data. The conecpt of SINDy is depicted in Fig. \ref{fig:Concept of SINDy}. We will consider the continuous-time nonlinear system as follows:

\begin{equation}
\begin{array}{l}
\dot{{\bm{x}}} = \bm{f}({\bm{x}}), \\
\end{array}
\label{eq:sindy11}
\end{equation}
where ${\bm{x}} \in \mathcal{X} \subset {\mathbb{R}}^m$ is the state and $\bm{f}: \mathcal{X} \rightarrow \mathcal{X} $ is the nonlinear function.

In SINDy approach, the nonlinear function can be represented as: 
\begin{equation}
\begin{array}{l}
{\bm{f}}_i \approx \bm{\Psi}_i(\bm{x})\bm{\Sigma}_i, \\
\end{array}
\label{eq:sindy21}
\end{equation}
where $\bm{\Psi}_i$ is a set of nonlinear candidate functions and $\bm{\Sigma}_i$ is a set of coefficient of candidate functions. 

To identify the governing equation of system, we take snapshots of state along $p$ times as
\begin{equation} 
\begin{split}
    \textbf{X} &= \left[ {\begin{array}{*{20}{c}} 
   \bm{x}_1 & \bm{x}_2 &  \cdots  & \bm{x}_p \end{array}} \right]^{\mathrm{T}}, \\
\end{split}
\label{eq:sindy3}
\end{equation}
where ${\textbf{X}} \in {\mathbb{R}}^{p \times m}$ is the snapshots of state.

We need the time derivative of state to employ SINDy. To obtain this value, numerical differential method such as simple forward
Euler finite-difference and total variation regularized derivative is used. The time derivative of state is also collected to make a snapshots as
\begin{equation} 
\begin{split}
    \dot{\textbf{X}} &= \left[ {\begin{array}{*{20}{c}} 
   \dot{\bm{x}}_1 & \dot{\bm{x}}_2 &  \cdots  & \dot{\bm{x}}_p \end{array}} \right]^{\mathrm{T}}, \\
\end{split}
\label{eq:sindy4}
\end{equation}
where $\dot{\textbf{X}}  \in {\mathbb{R}}^{p \times m}$ is the snapshots of time derivative of state.
 
Then, nonlinear candidate functions are designed to find a governing equation of system. Selecting the candidate function is recommended by a domain knowledge of system to enhance identification performance. If we do not have any knowledge, polynomial, trigonometric, and exponential functions are used to formulate the candidate function. The library of candidate function is 
\begin{equation} 
\begin{split}
    \bf{\Psi} (\textbf{X})  = \left[ {\begin{array}{*{20}{c}}
       \textbf{1} & \textbf{X} & \textbf{X}^2&  \cdots & \textbf{X}^d &  \cdots & sin(\bf{X})   & \cdots\\
    \end{array}} \right],
\end{split}
\label{eq:sindy6}
\end{equation}
where $d$ is the polynomial order. 

Therefore, the dynamical system can be represented in terms of the data matrices by:
\begin{equation}
\begin{array}{l}
\dot{\textbf{X}} \approx \bf{\Psi}(\textbf{X})\bf{\Sigma}, \\
\end{array}
\label{eq:sindy21}
\end{equation}

A penalty term is added to a regression problem to promote a sparsity of parameters. These penalty methods include $L_1$ (Lasso), $L_2$ (Ridge), and elastic net regularizations that have a $L_1$ and $L_2$ norm penalty to avoid an overfitting problem. In SINDy, the penalty term $L_1$ is used to obtain a parsimonious model for nonlinear candidate functions. The formulation of the sparse regression problem is 
\begin{equation} 
\begin{split}
    {\bf{\Sigma}}_{\text{k}} = \arg \min {\left\| \dot{\textbf{X}}_\text{k} - {\bf{\Psi}}(\textbf{X}){{\bf{\Sigma}}}_\text{k}    \right\|_2}^2 + \lambda {\left\| {\bf{\Sigma}}_{\text{k}} \right\|_1},
\end{split}
\label{eq:sindy423}
\end{equation}
where $\lambda$ is the regularizing parameter that makes a solution have a sparsity as a $L_1$ penalty and subscripts $\text{k}$ denote $k$th row. 
\subsection{Design of SINDy-based SOH Estimator}

\begin{figure}[h!]%
\centering
\includegraphics[width=75mm,clip]{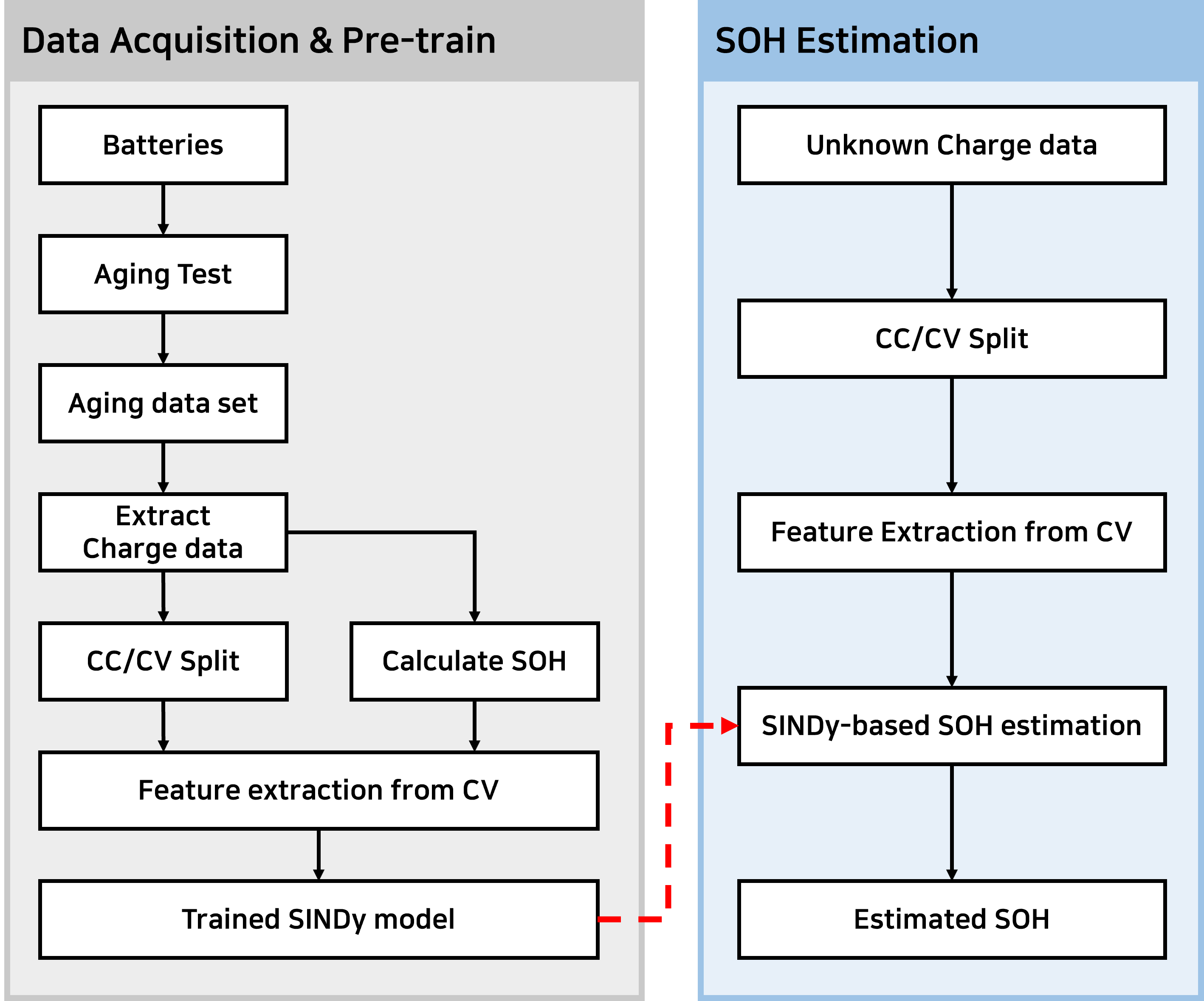}
  \caption{Concept of the SINDy-based SOH estimation.}
  \label{Concept}%
\end{figure}%
Figure \ref{Concept} shows the conceptual diagram of the SINDy-based SOH estimation method. First, training data are generated to train the model for the estimation of SOH. Since the training data consist of constant voltage charge data and CC-CV charge capacity as described above, two post-processing processes (CC/CV split, SOH calculation) are performed after the charge data selection. Second, the post-processed CV data are converted into data for singularity capture, which is utilized to train the SINDy model. Finally, the trained model estimates the SOH for the charging data (unknown), which are processed in the same way as the training data.
Based on the correlation analysis results in Section \ref{sec2}, state is defined by: 
\begin{equation} 
\begin{split}
\begin{array}{l}
 \bm{x} = {\left[ {\begin{array}{*{20}{c}}
 \mu, & \sigma, & Skew, & Kur, & \Delta I, & C_{\text{CV}}, & T \\
\end{array}} \right]^\mathrm{T}}. \\ 
 \end{array}
\end{split}
\label{eq:sindymracinput1}
\end{equation}

To estimate SOH, we assume that SOH can be represented as a discrete-time system 
\begin{equation} 
\begin{split}
\begin{array}{l}
 SOH_{k} = \bm{F}(\bm{x}_k), \\ 
 \end{array}
\end{split}
\label{eq:SOH1}
\end{equation}
where $\bm{F}$ is the nonlinear function of discrete-time system. 

Then, (\ref{eq:SOH1}) can be derived again using the nonlinear candidate function via 
\begin{equation} 
\begin{split}
\begin{array}{l}
 SOH_{k} = \bm{\Psi}(\bm{x}_k). \\ 
 \end{array}
\end{split}
\label{eq:SOH2}
\end{equation}

\noindent The polynomial function is only employed to formulate a library function. The order of polynomial function in a candidate function is set to 3 by try and error.  

In this paper, we use a sequential threshold least square algorithm (STLS), which is a hard threshold regression method. This method removes the parameter value if one parameter has a value smaller than the threshold. This property promotes a sparsity of data-driven model. Using the STLS, the sparse regression problem for SOH is 
\begin{equation} 
\begin{split}
    {\bf{\Sigma}} = \arg \min {\left\| \textbf{SOH} - {\bf{\Psi}}(\textbf{X}){{\bf{\Sigma}}}    \right\|_2}^2 + \lambda {\left\| {\bf{\Sigma}} \right\|_1},
\end{split}
\label{eq:sindy423}
\end{equation}
where $\textbf{SOH}$ is the snapshots of $SOH$. 

\section{Results and Discussion}
\label{sec4}
In this section, SOH estimation performances are verified using an evaluation metric such as mean absolute error (MAE), root mean square error (RMSE), and maximum error (MAX). The MAE and RMSE are used to evaluate overall performance, and MAX are used to evaluate the reliability of the system, which analyzes the performance in the worst-case scenario through the largest error value. The computation time is analyzed by a training time and a test time of each method. 

\subsection{SOH Model Training}
\begin{figure}[h]%
\centering
\includegraphics[width=85mm,clip]{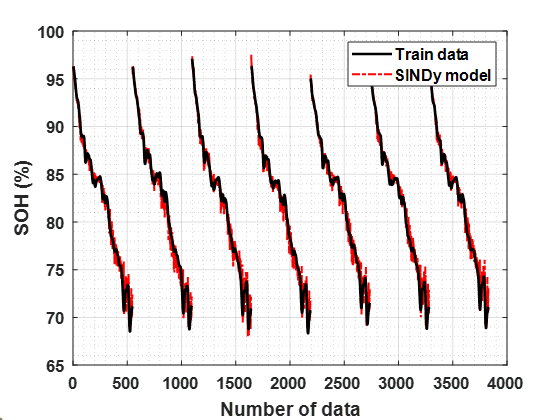}
  \caption{SOH model training result using SINDy.}
  \label{Train}%
\end{figure}

For the implementation and validation of the proposed method, Matlab 2024a was used, and the computer used consists of an AMD Ryzen Threadripper PRO 3975WX (CPU) and an NVIDA GeForce RTX 4060 (GPU). To train the SOH model, we used the data described in Section \ref{sec2}, with 7 LIBs used for training and 1 LIB used for validation. Figure \ref{Train} shows the training results of SINDy model. Through the SINDy algorithm, we were able to reduce the number of candidate functions that formulate the SOH model from 162 to 74 to obtain a sparsity of model. The trained model has an MAE of 0.5533, RMSE of 0.7531, and MAX of 5.1630, respectively.
 
\subsection{SOH Model Validation and Discussion}
\begin{figure}[h]%
\centering
\includegraphics[width=85mm,clip]{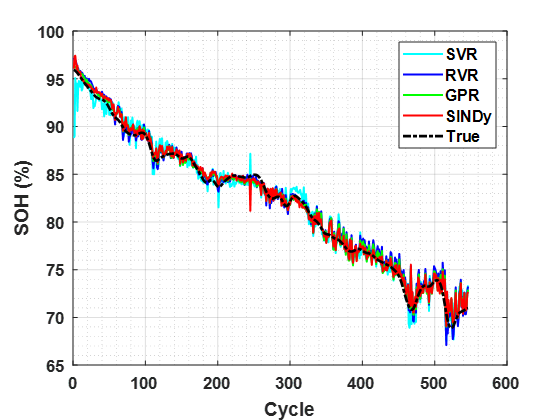}
  \caption{SOH estimation results using SVR, RVR, GPR and SINDy.}
  \label{Test01}%
\end{figure}

To validate an estimation performance, we perform the validation test using LIB data that were not used for training. The SINDy-based SOH estimator is compared with machine learning methods such as support vector regression, relevance vector regression, and Gaussian process regression with the same input data set to show fast estimation capability. Figure \ref{Test01} shows the estimation results with the proposed estimator and machine learning-based estimators. The SOH estimation performance with a proposed method and comparison methods is summarized in Table \ref{Test02}. 

\begin{center}
\begin{table}[h!]
\caption{Comparison of SOH estimation performance by model}
\centering
\begin{tabular}{|cc|cccc|}
\hline
\multicolumn{2}{|c|}{\multirow{2}{*}{}} & \multicolumn{4}{c|}{Methods} \\ \cline{3-6} 
\multicolumn{2}{|c|}{}     & \multicolumn{1}{c|}{SVR}    & \multicolumn{1}{c|}{RVR}    & \multicolumn{1}{c|}{GPR}     & SINDy  \\ \cline{3-6} \hline
\multicolumn{2}{|c|}{MAE}  & \multicolumn{1}{c|}{0.7483} & \multicolumn{1}{c|}{0.6577} & \multicolumn{1}{c|}{0.5798}  & \textbf{0.5516}  \\ \hline 
\multicolumn{2}{|c|}{MAX}  & \multicolumn{1}{c|}{7.1316} & \multicolumn{1}{c|}{4.5393} & \multicolumn{1}{c|}{ \textbf{4.3103}}  & 4.7101  \\ \hline 
\multicolumn{2}{|c|}{RMSE} & \multicolumn{1}{c|}{1.0413} & \multicolumn{1}{c|}{0.8875} & \multicolumn{1}{c|}{0.7759}  & \textbf{0.7636}  \\ \hline 
\end{tabular}
\label{Test02}
\end{table}
\end{center}

\begin{center}
\begin{table}[h!]
\caption{Comparison of computation time by model}
\centering
\begin{tabular}{|cc|cccc|}
\hline
\multicolumn{2}{|c|}{\multirow{2}{*}{}} & \multicolumn{4}{c|}{Methods} \\ \cline{3-6} 
\multicolumn{2}{|c|}{} & \multicolumn{1}{c|}{SVR}    & \multicolumn{1}{c|}{RVR}     & \multicolumn{1}{c|}{GPR}     & SINDy  \\ \hline
\multicolumn{2}{|c|}{\begin{tabular}[c]{@{}c@{}}Train Time (s)\end{tabular}} & \multicolumn{1}{c|}{0.9447}  & \multicolumn{1}{c|}{17.2233} & \multicolumn{1}{c|}{23.1552} & \textbf{0.1068}  \\ \hline
\multicolumn{2}{|c|}{\begin{tabular}[c]{@{}c@{}}Test Time (ms)\end{tabular}} & \multicolumn{1}{c|}{6.947}   & \multicolumn{1}{c|}{9.472}   & \multicolumn{1}{c|}{10.615}   & \textbf{0.094}   \\ \hline
\end{tabular}
\label{Test04}
\end{table}
\end{center}

The SOH estimation of SVR has a large SOH error of 7.1316$\%$ in the initial cycle and a bias in the overall cycle. The SOH estimation of GPR has a smaller RMSE and MAE than RVR, but the GPR performances are larger than the SINDy values. This is because SINDy constructs the model by explicitly reflecting the dynamical structure of the system, which enables it to learn complex relationships more accurately. 

SINDy has a relatively higher MAX performance at the 245 cycle than RVR and GPR because the SINDy can be obtained by minimizing the $L_2$ error norm to find a sparse model, so this characteristic can lead to large errors in certain local regions. However, in the process of estimating SOH, a measured capacity has a variation of about 0.2 Ah, which corresponds to approximately 8.3$\%$ variations of SOH. Therefore, the MAX of SINDy can be considered as a reasonable value for diagnosis of the health of system.

The calculation times for training and testing with a proposed method and the comparison methods are summarized in Table \ref{Test04}.
To train the SOH model, machine leraning-based methods need more time than SINDy-based method, and the GPR that have the smallest RMSE in comparison methods takes about 200 times calculation time compared to the SINDy-based estimator. The SINDy-based estimator has the shorter training and test time among all methods. In the test scenario, SINDy takes 70 to 100 times less calculation time than other methods because the machine learning methods based on the collected data require exponential computation time as the size of the data increases. However, the proposed method finds a governing equation of SOH model so that we can reduce the computation time without the collected data set. This is a significant advantage because it allows the system to estimate the SOH in real time. Therefore, the proposed method shows not only better SOH estimation performance but also fast computational speed than conventional machine learning-based methods.

\section{Conclusion}
\label{sec5}
In this paper, data-driven state-of-health estimation method for lithium ion batteries is proposed. The SINDy method is used to find a model of LIB degradation that is the result of an electrochemical reaction during charging and discharging. To select the input state, we perform a correlation analysis for statistical variables and domain-knowledge variables. 
Using SINDy and correlation analysis, we can improve the interpretability of the SOH system. It was found that the SINDy-based SOH estimation method is more computationally efficient and has the lowest RMSE and MAE than SVR, RVR, and GPR. In particular, the test time was about 100 times faster than other approaches. Future works involve integrating filtering methods such as extended Kalman filter and particle filter to reduce a noise effect. In addition, we plan to extend our research to SOH prediction to find a remaining useful life (RUL) of LIB system. 

\section*{Acknowledgment}
This research was supported by Unmanned Vehicles Core Technology Research and Development Program through the National Research  Foundation of Korea (NRF) and Unmanned Vehicle Advanced Research Center (UVARC) funded by the Ministry of Science and ICT, the Republic of Korea (2020M3C1C1A01083162).

\end{document}